\documentclass[conference]{IEEEtran}
\newcommand{\onedot}{\@addpunct{.}}
\newcommand{\our} {CRAFT Hand\xspace}
\newcommand{\ablation}{\texttt{LEAP}\xspace}
\newcommand{\ourshort}{CRAFT\xspace}
\def\@onedot{\ifx\@let@token.\else.\null\fi\xspace}
\DeclareRobustCommand\onedot{\futurelet\@let@token\@onedot}
\def\@onedot{\ifx\@let@token.\else.\null\fi}
\newcommand{\figref}[1]{Fig.~\ref{#1}}

\usepackage{xcolor}

\newcommand{\tabref}[1]{Tab.~\ref{#1}}

\usepackage{graphicx}
\usepackage{xspace}
\usepackage{caption}
\usepackage{array} 
\usepackage{booktabs}
\usepackage{tabularx}

\usepackage{float}
\usepackage{amsmath}
\usepackage{comment}
\let\labelindent\relax
\usepackage{enumitem}
\usepackage{times}

\usepackage[numbers]{natbib}
\usepackage{multicol}
\definecolor{mydarkblue}{rgb}{0,0.08,0.45}
\usepackage[bookmarks=true,hidelinks,colorlinks,linkcolor=mydarkblue,
    citecolor=mydarkblue,
    filecolor=mydarkblue,
    urlcolor=mydarkblue]{hyperref}

\begin{document}

% paper title
% \title{\our: A Compliant, Rigid, And Flexible Tendon-Driven Hand}
\title{\ourshort: A Tendon-Driven Hand with Hybrid Hard-Soft Compliance}

% You will get a Paper-ID when submitting a pdf file to the conference system

\author{
  Leo Lin$^{*1}$ \quad
  Shivansh Patel$^{*1}$ \quad
  Jay Moon$^{*1}$ \quad
  Svetlana Lazebnik$^{\dagger 1}$ \quad
  Unnat Jain$^{\dagger 2}$ \\[0.5em]
  \small{$^{1}$University of Illinois Urbana-Champaign \quad
  $^{2}$UC Irvine} \\[0.3em]
  \small{$^{*}$Equal contribution \quad $^{\dagger}$Equal advising}
}

\makeatletter
\let\@oldmaketitle\@maketitle
\renewcommand{\@maketitle}{\@oldmaketitle
\includegraphics[width=\textwidth]{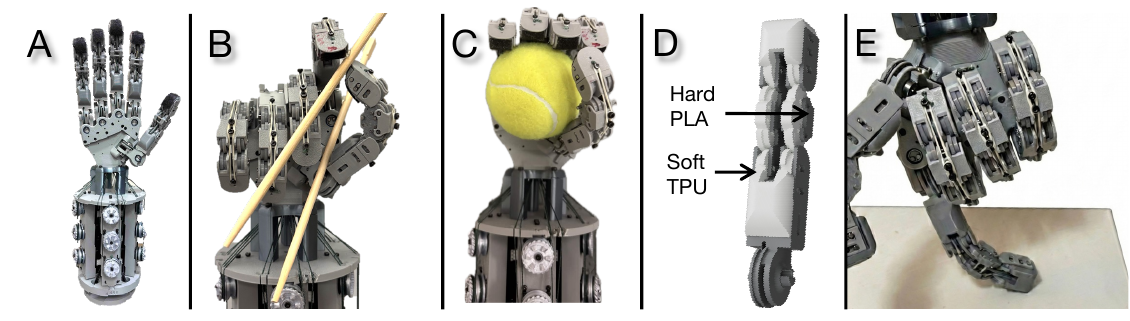}
    \captionof{figure}{\small \textbf{\ourshort Hand.} A) Full hand view showing the compact, anthropomorphic form factor with forearm-mounted actuators. B) \ourshort 
    holding chopsticks to show complex dexterity. C) \ourshort holding a tennis ball, demonstrating its human-like size. D) One finger composed of hard PLA links and 
    soft TPU joints. E) \ourshort conforming to a surface under 
    contact, illustrating passive compliance at the joints.}
  \label{fig:teaser}
  \vspace{-0.3in}
  \bigskip}

\makeatother
\maketitle
\thispagestyle{empty}
\pagestyle{empty}

\begin{abstract}
% We introduce \our, a tendon-driven anthropomorphic hand with hybrid hard-soft compliance that resolves the tradeoff between precision and robustness. Rigid hands offer repeatable kinematics but break under impact. Soft hands absorb contact forces but exhibit unpredictable, load-dependent behavior. Contact forces in manipulation are not uniform: impacts concentrate at joints, and structural loads concentrate in links. We place compliance where impacts happen, and use a rigid structure where forces must be transmitted. \ourshort implements this through rigid links and compliant joints with rolling contact surfaces that maintain repeatable motion paths. Benchmarking against the LEAP hand and structural analysis against RUKA and ORCA show that \ourshort matches rigid hand precision in a compact form factor. In teleoperation studies, passive compliance enables 100\% success on fragile objects where rigid hands fail. Evaluation on the Feix taxonomy confirms coverage of 33 grasp types from power to precision manipulation. The complete design costs \$600 USD, is assembled in under 5 hours from off-the-shelf parts, and will be released open-source.
We introduce \ourshort hand, a tendon-driven anthropomorphic hand with hybrid hard-soft compliance for contact-rich manipulation. The design is based on a simple idea: contact is not uniform across the hand. Impacts concentrate at joints, while links carry most of the load. \ourshort places soft material at joints and keeps links rigid, and uses rolling-contact joint surfaces to keep flexion on repeatable motion paths. Fifteen motors mounted on the fingers drive the hand through tendons, keeping the form factor compact and the fingers light. In structural tests, \ourshort improves strength and endurance while maintaining comparable repeatability. In teleoperation, \ourshort improves handling of fragile and low-friction items, and the hand covers 33/33 grasps in the Feix taxonomy. The full design costs under \$600 and will be released open-source with vision-based teleoperation and simulation integration. Project page: \url{http://craft-hand.github.io/}

% We propose \our, an low-cost, open-source dexterous anthropomorphic hand that bridges the gap between rigid and flexible soft robot hands. Existing hands often force a trade-off between precision and durability: rigid designs are precise but brittle, while soft designs are robust but lack durability and payload capacity. \ourshort hand resolves this through a hybrid architecture, combining rigid structural links with compliant, tendon-driven joints made from soft materials. This provides passive compliance to absorb extra forces without sacrificing the precise control. We validate the system through mechanical benchmarking against the rigid LEAP hand, showing that our design achieves comparable strength and repeatability. To demonstrate compatibility with modern learning frameworks, we implement a vision-based teleoperation pipeline that allows for intuitive data collection. In teleoperation tasks with delicate objects, the hand demonstrates a 100\% success rate, leveraging its passive compliance to safely manage contact forces. Additionally, the hand successfully executes 33 grasp types from popular grasp taxonomy, confirming its versatility. The complete hardware and software design is compatible with robot learning applications and will be open-sourced to support accessible research in robot manipulation.
\end{abstract}
\IEEEpeerreviewmaketitle

\section{Introduction}
Classical dexterous hands built around rigid linkages and direct-drive actuation have demonstrated precise, repeatable kinematics and strong actuation across decades of work. More recently, low-cost open platforms have made anthropomorphic dexterity substantially more accessible for robot learning research~\cite{shaw2023leap}. However, as the use of hands scales up, contact becomes the dominant failure mode: rigid structures transmit impacts directly into joints and linkages, and even minor collisions can interrupt long-running data collection. At its core, dexterous manipulation is about making and managing contact, which exposes a fundamental tension: a robot hand must be accurate enough to place contact where a learned policy expects it, yet robust enough to survive the inevitable collisions, slips, and suboptimal interactions that occur during data collection.

Soft robotic hands attempt to solve this by replacing rigid structures with compliant materials that deform under load~\cite{deimel2016novel,li2022brl}. This passive compliance is appealing: the hardware absorbs uncertainty that would otherwise require precise control~\cite{McGeer01041990}. But compliance introduces a different problem. Without a rigid structure, soft hands have limited load capacity and exhibit configuration-dependent kinematics. A soft finger can bend differently depending on how much it is already supporting, making its motion and force transmission difficult to model and repeat.

Both failure modes, rigid hands that break under contact and soft hands that lose geometric structure, stem from applying uniform material choices across the hand. In manipulation, impacts and conforming contact concentrate at joints, where the hand changes direction and collides with objects. The segments between joints, by contrast, must reliably transmit actuation forces from the motors, so they must be rigid. This spatial separation motivates a hybrid style of compliance: \emph{place compliance at joints where impacts and contact uncertainty concentrate, and use rigid links where geometric predictability and load paths matter}. The goal is not ``soft everywhere,'' but ``soft where it helps,'' so passive deformation absorbs shocks and regulates forces without sacrificing kinematic structure.

To this end, we propose \ourshort hand (\figref{fig:teaser}), a tendon-driven anthropomorphic hand with hybrid hard-soft compliance that costs under \$600 and weighs 800g. The hand is fully 3D-printed:  finger segments use rigid polylactic acid (PLA) material, while the finger joints use soft thermoplastic polyurethane (TPU), as highlighted in \figref{fig:teaser} D. Introducing soft joints raises two practical challenges. First, soft joints are difficult to control predictably: in conventional tendon-driven fingers, a single tendon pulls through both the proximal interphalangeal (PIP) and distal interphalangeal (DIP) joints, causing one joint to bend more than the other depending on load and posture. We address this with a bidirectional mechanical linkage that couples the PIP and DIP joints, so both bend equally and simultaneously regardless of external load, producing repeatable finger postures. Second, soft flexure joints accumulate fatigue under repeated bending, eventually breaking. We address this with rolling-contact joints at the PIP and DIP locations: rather than bending at a fixed point, the joint rotates along a constrained curved interface, distributing load across the TPU surface and significantly reducing stress concentration~\cite{ahn2023asymmetric}. Together, these two mechanisms allow the compliant joints to remain both predictable and durable through contact-rich operation.

Actuation comes from 15 actuators located behind the wrist, with three actuators per finger driving flexion and side-to-side motion through tendons. Routing actuation away from the fingers keeps the hand compact and anthropomorphic in form factor. It also removes the actuators from contact zones, reducing the risk of damage during collisions, and keeps the fingers light, which simplifies full-system teleoperation. Tendon routing also provides mechanical advantage, improving holding strength and reducing effort during sustained grasps. The hand is modular and fully 3D-printable, so damaged components can be replaced quickly without disassembling the system.

Because the hand is intended for data-driven learning, we pair the hardware with tools for data collection and reinforcement learning. For teleoperation, we use a single RGB camera with vision-based whole-body tracking to control both the arm and hand simultaneously, enabling natural data collection without wrist-mounted sensors or gloves. The compliant joints reduce the operator's cognitive load during collection: passive contact absorption means demonstrations remain valid even when finger placement is imprecise, which is particularly important for fragile or deformable objects. We also release URDF and MuJoCo XML models that preserve the hand's kinematics while using practical approximations for the soft joints, suitable for reinforcement learning.

We evaluate three questions. First, does our hybrid compliance preserve structural performance? Benchmarked against the rigid LEAP hand~\cite{shaw2023leap}, \ourshort matches strength, precision, and repeatability, while giving the benefits of compliance. Second, does compliance enable more robust manipulation? Teleoperation across five tasks, from rigid objects to fragile and low-friction items, shows \ourshort achieves consistently higher success by passively regulating contact forces that would otherwise demand precise control. Third, does the hybrid design support general dexterity? We validate against the Feix grasp taxonomy~\cite{feix2015grasp}, demonstrating success on all 33/33 grasps from power grips to precision pinches.

In summary, our contributions are as follows:
\begin{itemize}[left=0pt, labelsep=5pt]
\item \textbf{Hybrid hard-soft hand}: \ourshort localizes compliance at joints and rigidity at links, resolving the precision-robustness tradeoff at under \$600, fully open-source.
\item \textbf{Coupled rolling-contact joints}: A bidirectional PIP-DIP linkage and rolling-contact surfaces produce predictable, repeatable joint motion while absorbing impact forces.
\item \textbf{Structural benchmarking}: \ourshort matches the LEAP hand~\cite{shaw2023leap} in repeatability and pull-out strength while surviving impacts that would break the rigid baseline.
\item \textbf{Manipulation and dexterity validation}: 100\% success on fragile objects in teleoperation studies, and full coverage of all 33 Feix grasp types.
\end{itemize}

\section{Related Work}
\label{sec:related}
\vspace{0.05in}

\noindent\textbf{Direct-Drive Robotic Hands} Historically, dexterous manipulation research has relied on fully actuated, direct-drive hands~\cite{allegrohand,barretthand,dlrhithand2,bhirangi2022all}. These systems employ rigid linkages with motors at joints or in the palm for high torque and precise control. The DLR-HIT Hand II~\cite{dlrhithand2} integrates motors into finger bases and phalanges for a compact design. However, it comes at steep costs (\$15,000–\$80,000) and has kinematic limitations. Meanwhile, the Allegro Hand~\cite{allegrohand}, while robust, lacks abduction/adduction degree of freedom (DoF), restricting its ability to perform human-like opposition and power grasps.

Recent efforts focus on bridging this gap through affordable, open-source hardware (approx. \$2,000). The LEAP Hand~\cite{shaw2023leap} addresses kinematic shortcomings of direct-drive predecessors; its universal abduction-adduction mechanism eliminates workspace dead zones. Similarly, the D'Manus~\cite{bhirangi2022all} offers affordable dexterity. However, housing motors within fingers or palms increases thickness, creating a bulky form factor exceeding human dimensions. Furthermore, these rigid hands rely on complex active impedance control rather than mechanical compliance to manage contact forces. Consequently, inaccuracies cause high-force collisions, risking damage. In contrast, our hand integrates a hybrid soft-rigid structure with tendon-driven compliance, adapting to contact while retaining a human-like form factor.

\vspace{2mm}
\noindent\textbf{Tendon-driven and compact designs} While direct-drive hands offer simplicity and robustness, placing motors within the fingers or palm creates bulky, non-anthropomorphic form factors. To achieve the compact size of a human hand, researchers use tendons to relocate motors to the forearm~\cite{shadowhand,jacobsen1986design,inmoov_hand,park2020open,liu2008multisensory,zorin2025ruka,christoph2025orca}. The Shadow Hand~\cite{shadowhand}, a long-standing benchmark in this category, uses a complex array of tendons to actuate 24 joints, mimicking human kinematics. However, its high cost and closed-source nature have limited its use in large-scale learning experiments.

Recent open-source initiatives have sought to democratize this morphology. The RUKA hand~\cite{zorin2025ruka} achieves a 15-DoF anthropomorphic design, more compact than direct-drive alternatives by housing servos in a forearm unit. Similarly, the ORCA Hand~\cite{christoph2025orca} addresses tendon fragility via "popping" joints that dislocate under stress rather than breaking, ensuring reliability for operation. Despite these advancements, most tendon-driven hands remain rigid. Stiff tendons transfer motor torque directly to joints, meaning these systems still rely on active impedance control rather than passive compliance to manage interaction forces. Our hand utilizes soft materials, providing inherent compliance that allows fingers to naturally conform without complex control.

\vspace{2mm}
\noindent\textbf{Compliance and Soft Robotics}
To overcome the challenges posed by the stiffness of rigid and tendon-driven mechanisms, where imperfect state estimation can lead to damaging collisions, researchers have turned to passive compliance as a hardware-level solution. Early efforts~\cite{ma2017yale,zisimatos2015create,deimel2016novel,park2020open,mannam2023framework} utilized elastomer flexure joints to achieve this, allowing fingers to conform passively without complex control. However, these designs are typically underactuated (low DoF), limiting their ability to perform complex in-hand manipulation.

To combine robustness with dexterity, recent works~\cite{toshimitsu2023getting,li2022brl} have explored hybrid soft-rigid architectures. The closest to our work is Leap V2~\cite{shaw2025leap}, which successfully integrates elastomer joints into a high-DoF (22 joints) system. However, while mechanically capable, the Leap V2 remains expensive (\$5000) and relatively bulky. In contrast, our hand achieves a significantly more compact, anthropomorphic form factor at a fraction of the cost (\$600), without sacrificing the high-DoF compliance essential for robust manipulation. Furthermore, our use of rolling contact joints improves durability compared to flexure-based joints, which break under repeated use.

\section{Kinematic Design}
\label{sec:kinematic}
This section presents the mechanical implementation of our hybrid design principle, describing the hand architecture, joint mechanisms, and tendon routing.

\begin{figure}[t]
\vspace{-0.3cm}
  \centering
  \includegraphics[width=0.8\linewidth]{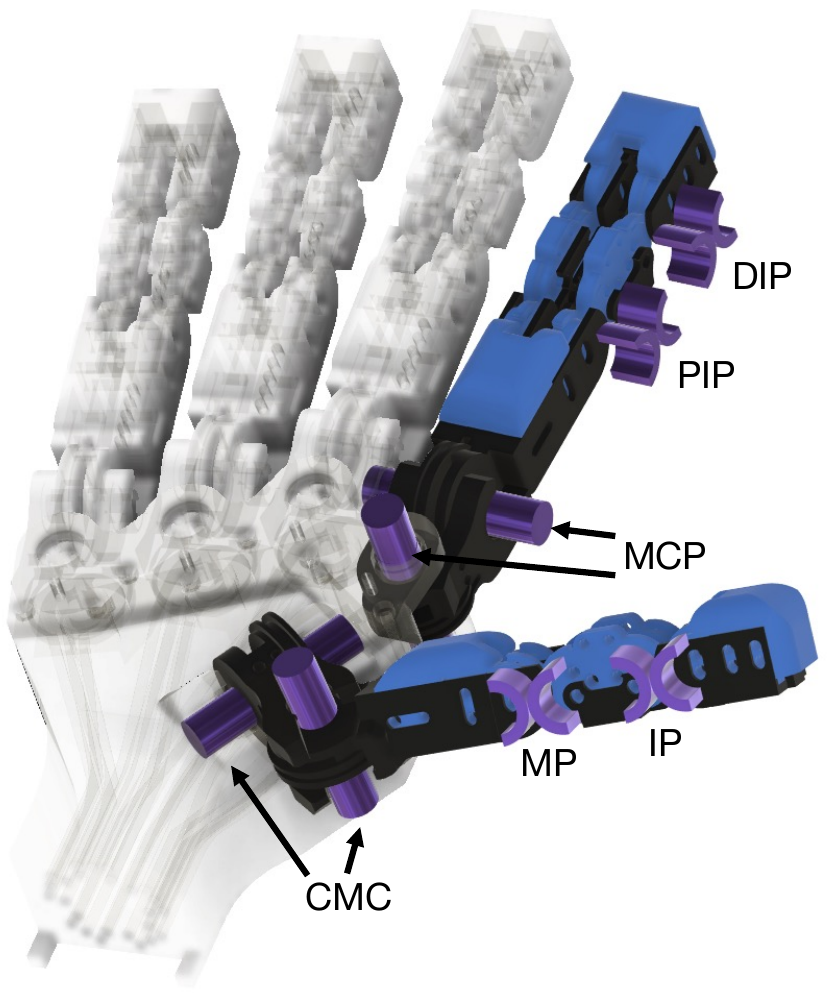}
  \vspace{-0.1cm}
    \caption{\small \textbf{Joints Structure.} Black denotes the rigid components, while blue denotes the compliant (TPU) components. The index finger and thumb are highlighted to illustrate kinematics, with purple components indicating joint axes. Fingers feature a 2-DoF MCP and coupled PIP/DIP joints; the thumb mirrors this with a 2-DoF CMC and coupled MP/IP joints. PIP/DIP and MP/IP employ rolling contacts, while MCP/CMC use the cylindrical snap-fit interface.}
    \vspace{-0.5cm}
  \label{fig:joints}
\end{figure}

\vspace{2mm}
\noindent\textbf{\ourshort Hand Structure:} The entire hand structure is shown in \figref{fig:teaser} (top-right) and an X-ray view of \ourshort is shown in \figref{fig:joints}. To maintain anthropomorphism, \ourshort has 95 mm palm and 103 mm finger length, totaling 198 mm. These dimensions align with median male hand measurements~\cite{greiner1991hand}, ensuring the system is compact for teleoperation. \ourshort weighs 800g and employs 15 actuators for 15 active and 5 passive degrees of freedom (DoF). To maintain compactness, all motors are located in the forearm.

\begin{figure}[b]
\vspace{-0.5cm}
  \centering
  \includegraphics[width=\linewidth]{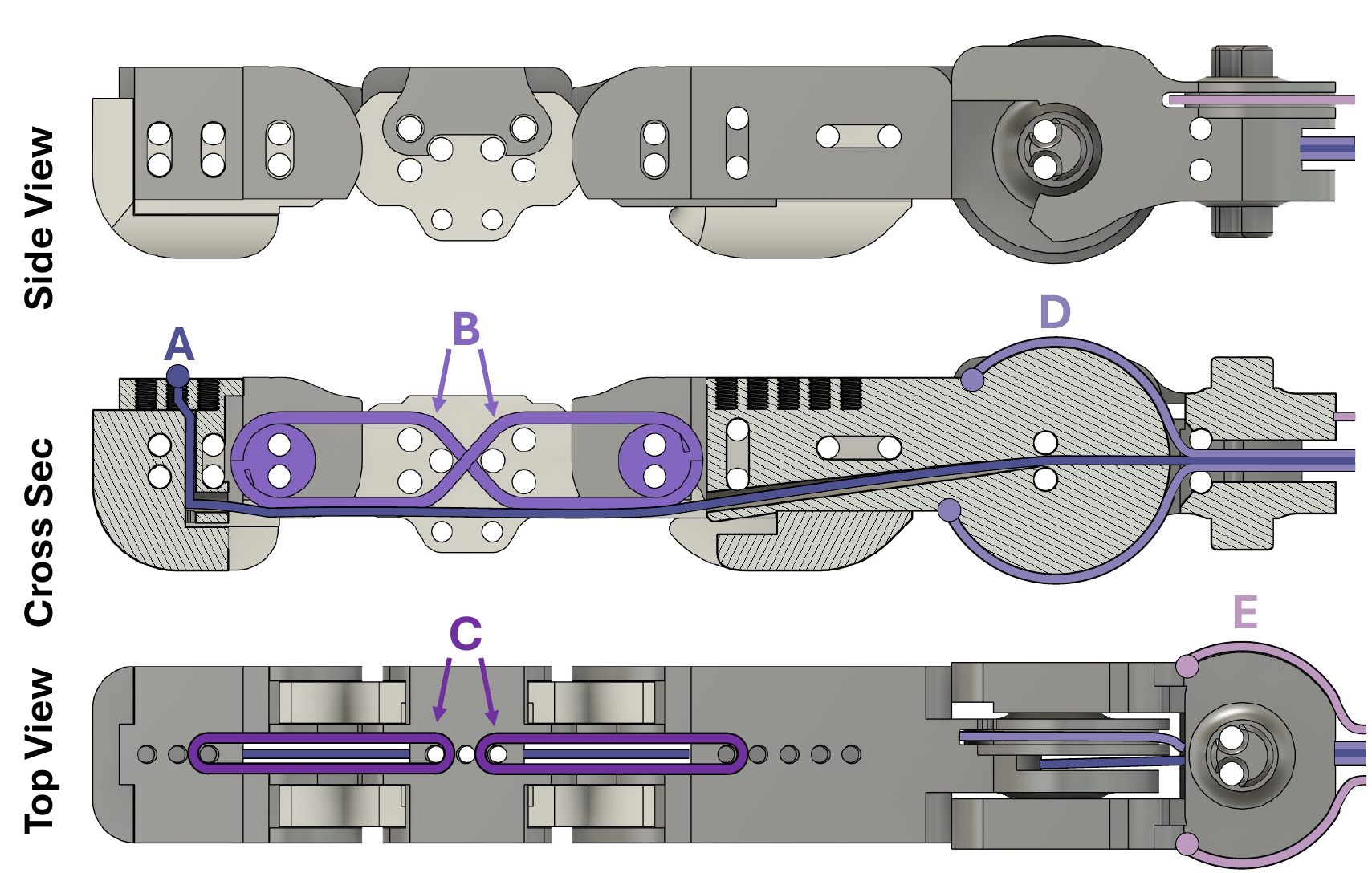}
  \vspace{-0.6cm}  % tighten space above caption (adjust as needed)
    \caption{\textbf{Finger Structure.} Dark gray denotes the hard (PLA) components and light gray denotes the soft (TPU) components. (A) PIP/DIP Tendon. (B) Bidirectional Linkage. (C) Elastic bands. (D) MCP Flexion/Extension Tendons. (E) Abduct/Adduct Tendons.}
  \label{fig:finger}
\end{figure}

\vspace{2mm}
\noindent\textbf{Finger and Joints Structure:} 
We denote the finger joints using standard anatomy: the metacarpophalangeal (MCP) joint at the finger base, the proximal interphalangeal (PIP) joint in the middle, and the distal interphalangeal (DIP) joint closest to the fingertip, as shown in \figref{fig:joints}. We assign three actuators to each finger to control \emph{(i) coupled flexion of the PIP and DIP joints}, \emph{(ii) abduction/adduction of the MCP joint}, and \emph{(iii) flexion/extension of the MCP joint}.

Each finger comprises distinct rigid PLA and soft TPU parts, shown in \figref{fig:finger}. Rigid PLA wraps the segments to maintain alignment and transmit tendon forces, while the PIP and DIP joints consist entirely of soft TPU, allowing elastic deformation under impact.

At the PIP and DIP joints, a single tendon (\figref{fig:finger}\textbf{A}) drives both joints through a bidirectional TPU linkage (\figref{fig:finger}\textbf{B}) that couples them so motion at one drives the other with equal rotation. \ourshort uses rolling contact joints~\cite{ahn2023asymmetric} at these locations rather than the pin joints~\cite{shaw2023leap,zorin2025ruka} common in soft robotic fingers. Two circular surfaces roll relative to each other, distributing load across the joint surface and eliminating the stress concentrations that cause flexure joints to fracture under repeated use. This geometry constrains motion to a well-defined kinematic path regardless of load, so both joints follow the same trajectory on every cycle. Together, the coupling linkage and rolling-contact geometry keep finger posture repeatable and predictable, allowing accurate modeling. Passive elastic bands (\figref{fig:finger}\textbf{C}) return both joints to neutral when tendon tension is released.

At the MCP joint, a cylindrical snap-fit interface motivated by Christoph \textit{et al.}~\cite{christoph2025orca} simplifies assembly and pops out under excessive force, reducing the risk of joint breakage. A flexion/extension tendon (\figref{fig:finger}\textbf{D}) and an abduction/adduction tendon (\figref{fig:finger}\textbf{E}) actuate the MCP independently, and both are backdrivable, allowing external forces to drive the transmission in reverse. Metal dowels throughout the finger provide structural support and low-friction guidance for all tendons.

The modular design allows a full finger to be swapped by detaching the dowel pins, without disassembling the hand or rewiring tendons. The thumb is mounted laterally to replicate human anatomy. Its joints use distinct anatomical names, CMC, MP, and IP, but the mechanical implementation mirrors the fingers exactly: the CMC uses the same snap-fit interface for 2-DoF motion, and the MP and IP joints use the same rolling-contact design and coupled-linkage mechanism as the finger PIP and DIP joints.

\vspace{2mm}
The routing layout for a single finger is shown in \figref{fig:finger}. Two opposing pairs control MCP flexion/extension and abduction/adduction, while a single tendon drives coupled PIP/DIP flexion. We use high-strength braided line to withstand substantial tension. To minimize friction, metal dowel pins serve as smooth guide surfaces. Our routing layout is designed for visual and mechanical simplicity, allowing users to readily identify tendons during maintenance. The motors are arranged in a compact pentagonal shape, with each face serving a single finger. Similar to Christoph \textit{et al.}~\cite{christoph2025orca}, we use a ratchet spool to quickly re-tension the tendons.

\section{Teleoperation and Simulation Support}
\label{subsec:whole_arm_teleop}
For a hand to be useful for robot learning, it must map reliably to both a human operator and a simulator. We address this in two steps: first, a calibration procedure that aligns the operator's hand range to the robot's joint limits, and then integration with vision-based teleoperation for data collection and simulation for reinforcement learning.

\vspace{2mm}
\noindent\textbf{Teleoperation.} To teleoperate the hand, we must know how many motor turns move each joint across its full range. Following ~\cite{christoph2025orca}, we employ a joint-wise calibration procedure to align each operator's movements with \ourshort hardware. We begin by manually moving each robot joint to its extremes to record its range. We then record the operator's biological limits using HaMeR~\cite{pavlakos2024reconstructing} as they move their fingers through their full range of motion, covering both flexion/extension and abduction/adduction. During teleoperation, a linear function maps the operator's current joint position, normalized against their biological range, to the corresponding angle within the robot's calibrated limits, ensuring accurate retargeting across all degrees of freedom.

\begin{figure}[t]
  \centering
  \includegraphics[width=1\linewidth]{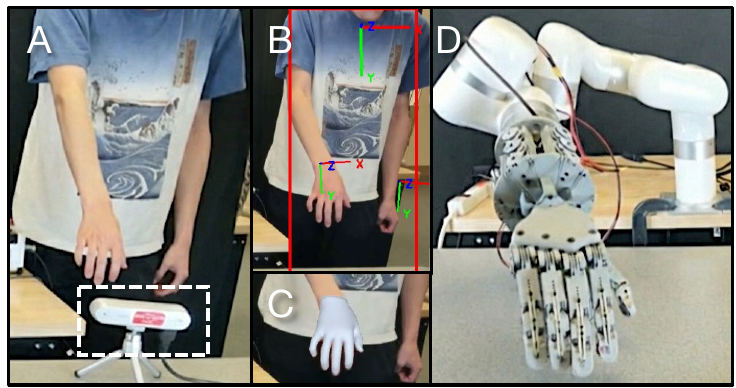}
  \vspace{-0.5cm}  % tighten space above caption (adjust as needed)
    \caption{\small \textbf{Whole-Arm Teleoperation.} \textbf{(A)} An operator stands in front of a single RGB camera (highlighted with a dashed box). \textbf{(B)} FrankMocap estimates the 3D pose of the operator's wrist relative to their torso, used for whole-arm retargeting. \textbf{(C)} HaMeR estimates the operator's hand pose for finger retargeting. \textbf{(D)} The estimated poses are mapped to \ourshort, mounted as the end-effector of a robotic arm.}
  \label{fig:vision-teleop}
  \vspace{-4mm}
\end{figure}

    % \caption{\small \textbf{Whole-Arm Teleoperation and Simulation.} Left: An operator controls the robot arm and hand using a single RGB camera (highlighted with a dashed box) placed in front of the operator. Middle: Operator performing teleoperation with hand pose overlay. Right: \ourshort model in MuJoCo simulation environment.}

With calibration in place, we follow the vision-based teleoperation framework of Sivakumar \textit{et al.}~\cite{sivakumar2022robotictelekinesislearningrobotic} to collect demonstration data. As shown in \figref{fig:vision-teleop}\textbf{A}, the operator stands in front of a single RGB camera. FrankMocap~\cite{rong2020frankmocap} estimates the 3D pose of the operator's wrist relative to their torso (\figref{fig:vision-teleop}\textbf{B}), which is mapped to the robot's workspace for whole-arm retargeting. Simultaneously, HaMeR estimates the operator's hand pose (\figref{fig:vision-teleop}\textbf{C}) for finger retargeting. An exponential moving average is applied to smooth targets and mitigate noise inherent to vision-based tracking. Together, these estimates drive \ourshort, mounted as the end-effector of a robotic arm (\figref{fig:vision-teleop}\textbf{D}), enabling simultaneous control of the arm and hand with no wrist-mounted sensors or gloves.

\begin{figure}[h]
  \centering
  \includegraphics[width=0.6\linewidth]{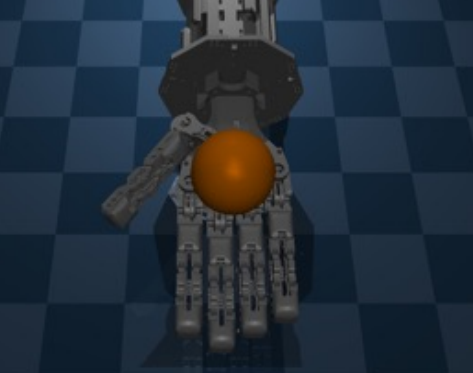}
  % \vspace{-0.5cm}  % tighten space above caption (adjust as needed)
    \caption{\small \textbf{\ourshort in MuJoCo Simulation.} \ourshort grasping 
a sphere, with rolling contact joints modeled as equality constraints and 
joints treated as revolute constraints.}
  \label{fig:sim}
  \vspace{-4mm}
\end{figure}

\vspace{2mm}
\noindent\textbf{Simulation.} For simulation-based policy learning, we provide URDF~\cite{makoviychuk2021isaac} and XML~\cite{todorov2012mujoco} files compatible with widely adopted simulators~\cite{andrychowicz2020learning,qi2023hand,chen2023visual}. Since the physical behavior of soft materials does not transfer reliably from simulation to the real world, we avoid soft-body physics and instead model the rolling-contact joints using equality constraints that enforce surface contact conditions and treat them as revolute joints, as shown in \figref{fig:sim}. This approximation maintains computational efficiency while preserving the essential kinematics for policy learning.
\section{Experiments}
\label{sec:experiments}
We evaluate \ourshort along three axes. First, we test structural performance to measure the strength and precision that the hybrid design achieves relative to a rigid baseline. Second, we assess manipulation performance to determine whether passive compliance reduces operator effort and failures during teleoperation. Third, we validate grasp versatility to confirm that the hybrid design maintains the kinematic range and strength required for general manipulation.

\subsection{Structural Tests}
To evaluate the structural robustness of \ourshort hand, we conduct three assessments. We evaluate the strength, precision, and endurance of CRAFT to the LEAP hand~\cite{shaw2023leap} using standardized structural tests, consistent with prior work~\cite{zorin2025ruka,christoph2025orca}. 
It is a widely used rigid, direct-drive hand. To ensure a fair comparison, we assembled the LEAP hand using the same motors as \ourshort hand (Dynamixel XL330-M288-T instead of the $4\times$ expensive Dynamixel XC330-M288-T), hereafter denoted as \ablation. This allows us to make a head-on comparison between \ourshort and a rigid, direct-drive linkage (\ablation).

\vspace{2mm}
\noindent\textbf{Pull-out Test (Strength).}
This experiment quantifies the maximum payload a finger can retain before mechanical failure. We command the finger to a fully flexed position and apply an external force opposing the flexion direction. The failure threshold is defined as the point where the finger slips or deflects by more than $15^\circ$ from the target pose. We track the deflection angle using ArUco markers attached to the finger.
As detailed in \tabref{tab:pullout}, \ourshort withstands 15.29\,N of pull-out force, nearly double the 8.67\,N measured for the \ablation hand. This performance gap highlights the mechanical advantage of the tendon routing system. By distributing the load through the tendon, \ourshort reduces the direct torque demand on the motors compared to the direct-drive linkage of the \ablation, allowing for higher holding forces with identical actuation hardware.

\begin{table}[h]
    \centering
    \footnotesize
    % Redefine X columns to vertically center content (like 'm' columns)
    \renewcommand{\tabularxcolumn}[1]{m{#1}}
    
    % X columns automatically fill the remaining space. 
    % @{} removes the gap between the two image columns.
    \begin{tabularx}{\linewidth}{l >{\centering\arraybackslash}X@{} >{\centering\arraybackslash}X}
    \toprule
    \textbf{Hand} & \ourshort & \ablation \\ \midrule
    \textbf{Visual} & 
    % \linewidth in an X column refers to the cell width, not the page width
    \includegraphics[width=\linewidth]{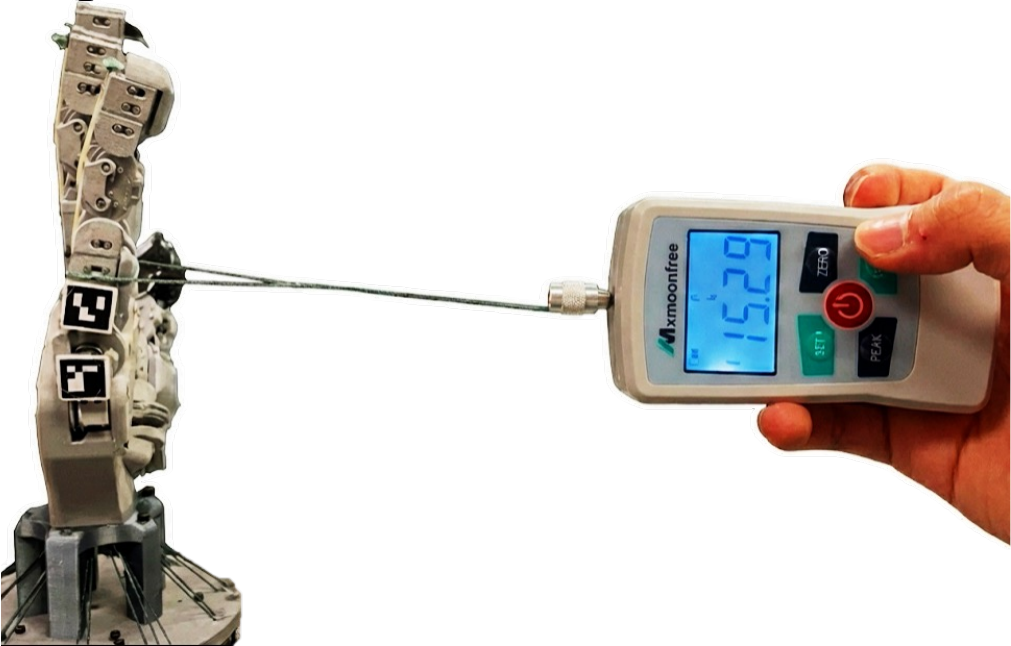} & 
    \includegraphics[width=\linewidth]{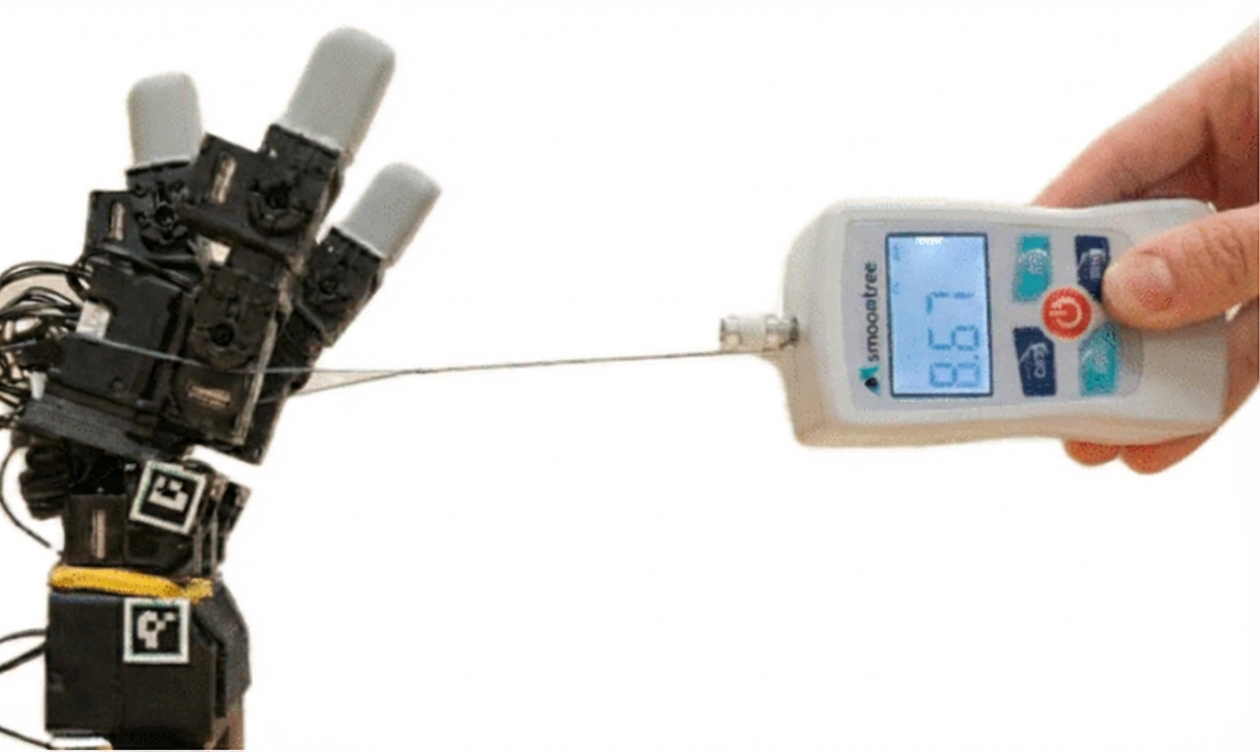} \\ \addlinespace
    \textbf{Strength} & 15.29 & 8.67 \\ 
    \bottomrule
    \end{tabularx}
    \caption{\small \textbf{Pull-out Strength Comparison.} We measure the resistance of a flexed finger against an opposing external force (in N). The maximum force is recorded immediately prior to a fingertip deflection exceeding $15^\circ$. \ourshort demonstrates significantly higher retention strength due to the mechanical advantage of its tendon.}
    \label{tab:pullout}
    \vspace{-0.5cm}
\end{table}

\begin{figure}[h]
  \centering
  \includegraphics[width=\linewidth]{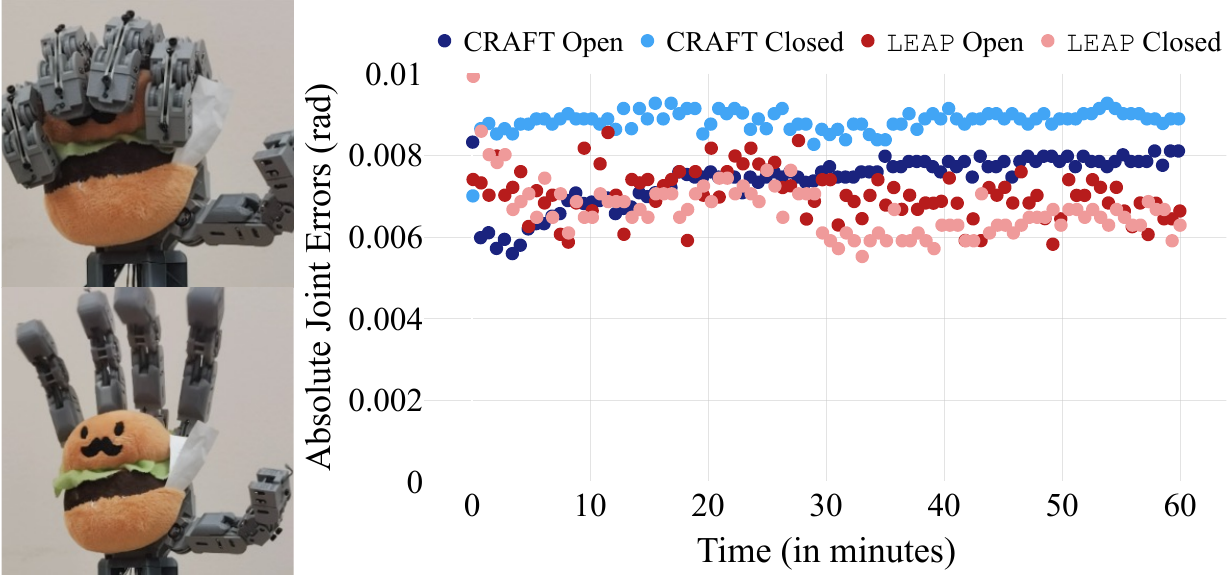}
  \vspace{-0.3cm}  % tighten space above caption (adjust as needed)
    \caption{\small \textbf{Repeatability under Cyclic Load.} \textit{Left:} \ourshort performing continuous grasp-release cycles on an object. \textit{Right:} Joint angle tracking error over one hour of operation. Despite the use of flexible tendons, \ourshort (blue dots) maintains a tracking error of $<0.01$\,rad, comparable to the rigid \ablation baseline (red dots), demonstrating consistent control. (shades depict joints closing and opening)}
  \label{fig:repeatability}
  \vspace{-0.2cm}
\end{figure}

\vspace{2mm}
\noindent\textbf{Repeatability Test (Precision).}
We investigate whether the tendon-driven mechanism introduces position errors or slack over prolonged operation. We program both hands to perform a grasp-and-release sequence on a plush toy continuously for one hour. \figref{fig:repeatability} presents the joint angle tracking error over the duration of the experiment. The error is calculated using the motor encoder values. Both \ourshort and \ablation maintain consistent performance with mean tracking errors remaining below 0.01\,rad. This result demonstrates that \ourshort achieves consistency similar to rigid hands, confirming that the benefits of compliance do not come at the cost of repeatability.

\begin{figure}[h]
     \centering
     \vspace{-0.1in}
     \includegraphics[width=0.95\linewidth]{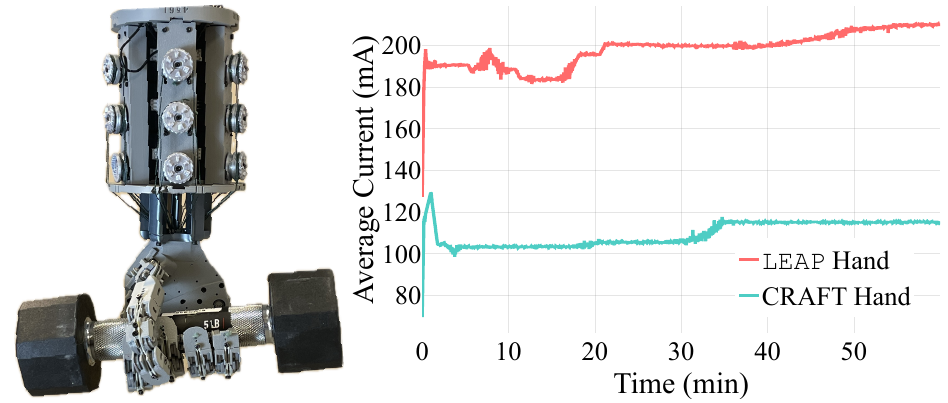}
     % \vspace{-0.1cm}
    \caption{\small \textbf{Holding Test}. We make \ourshort and \ablation hand grasp a heavy 5lb weight in its palm for one hour.  On the vertical axis, we show that the average current running through \our is two times less compared to \ablation.  The right axis shows that the current use initially increases. However, it still holds the weight and uses around one-third of its maximum possible current of 600mA.}
     \label{fig:endurance}
\end{figure}

\noindent\textbf{Holding Test (Endurance).}
Finally, we evaluate the hand's efficiency during high-load static holding. Both \ourshort and \ablation are commanded to grasp and hold a 5\,lb dumbbell vertically for one hour (\figref{fig:endurance}). We monitor the current draw to assess motor strain and thermal throttling. The results show that the average current consumption of \ourshort is approximately 50\% lower than that of \ablation. While current draw in both hands rises initially as motor temperature increases, \ourshort stabilizes well below the motor's 600\,mA limit. The friction inherent in the tendon routing acts as a passive braking mechanism, assisting in load retention and reducing the active torque required from the motors. In contrast, the direct-drive \ablation must constantly supply higher torque to combat the moment arm of the weight, leading to higher power consumption and thermal stress.

\subsection{Teleoperation Tests}
\label{subsec:teleop}

\begin{figure}[t]
  \centering
  \includegraphics[width=0.5\linewidth]{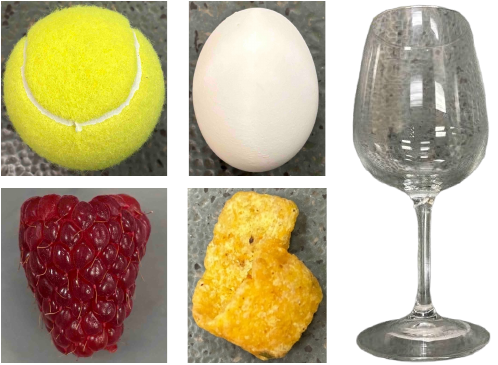}
  \vspace{-0.1cm}  % tighten space above caption (adjust as needed)
    \caption{\small \textbf{Teleoperation User Study Objects.} To validate real-world utility, we evaluate teleoperation performance on five objects: (1) a ball (rigid grasping), (2) a wine glass (large and fragile), (3) an egg (low friction and fragile), (4) a raspberry (small and deformable), and (5) a frito chip (extremely delicate).}
  \label{fig:exp_objs}
  \vspace{-0.5cm}
\end{figure}

To validate whether \ourshort's compliance translates into improved utility in real-world manipulation, we conduct a teleoperation user study comparing \ourshort against the rigid \ablation baseline. We hypothesize that the passive compliance provided by the soft components will allow for higher success rates with fragile objects and faster completion times by reducing the cognitive load required for precise alignment. To test this, we experiment on five objects shown in \figref{fig:exp_objs}, each selected to isolate a specific manipulation challenge: 
\begin{enumerate}[leftmargin=*]
    \item \textbf{Grasping a Ball:} Requires no compliance, as the ball is solid. This is a reasonably easy object to grasp.
    \item \textbf{Lifting a Wine Glass:} Requires handling a large, fragile object susceptible to shattering under excessive force.
    \item \textbf{Handling an Egg:} Requires managing low-friction fragility, demanding a grasp that is delicate enough to prevent crushing but firm enough to prevent slip.
    \item \textbf{Picking a Raspberry:} Requires extreme delicacy at a small scale, where the object is easily deformed by minor force.
    \item \textbf{Lifting a Frito Chip:} Requires extreme delicacy, serving as a test for passive compliance.
\end{enumerate}

\begin{figure}[t] 
  \centering
  \includegraphics[width=1\linewidth]{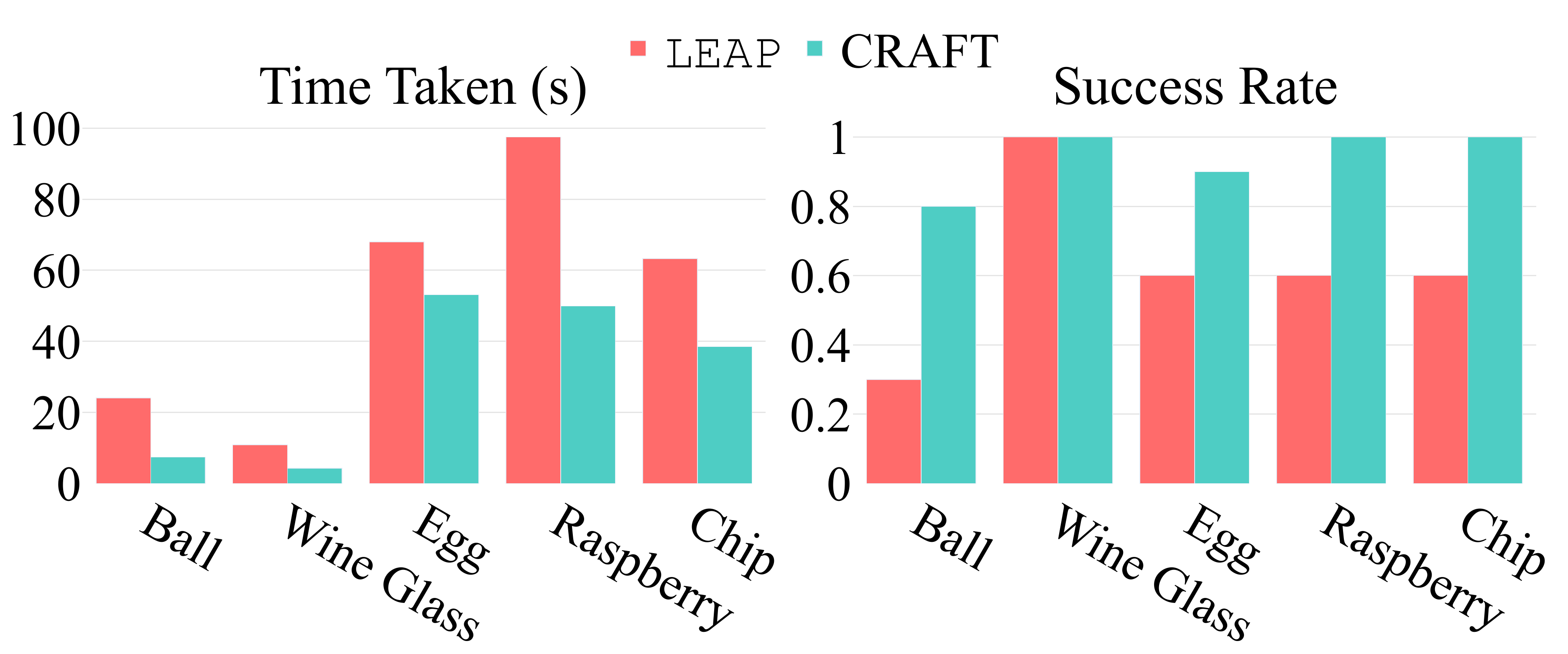}
  \caption{\small \textbf{Teleoperation User Study Results.} Comparison of average completion time and success rate for five manipulation tasks. \ourshort consistently outperforms the LEAP hand, particularly in handling fragile objects (Raspberry, Chip) where compliance prevents damage, and in dynamic grasping (Ball) where adaptation simplifies capture.}
  \label{fig:teleop_results}
  \vspace{-0.5cm}
\end{figure}

To strictly evaluate the grasping capability of the hand, the robot arm is fixed at an optimal pre-grasp pose for each task. Users control only the hand's fingers via teleoperation. Prior to testing, participants were briefed on the mechanical differences between the hands and given two minutes to familiarize themselves with the teleoperation interface. A limit of 150 seconds per trial was enforced. We run 10 trials for each hand with different people. We plot the results in \figref{fig:teleop_results} and discuss them below.

We measure both success rate and completion time across trials. Success rate captures whether the hand can complete the task at all, while completion time reflects the cognitive and control effort required from the operator. We run 10 trials for each hand with different people. We plot the results in \figref{fig:teleop_results} and discuss them next.

We first examine the grasping ball task to test performance on rigid, non-deformable objects. The bulky structure of \ablation proved a confounding factor, making effective grasps difficult and resulting in taking a longer time and achieving a lower success rate. Since the ball is rigid, \ourshort's passive conformity played a minor role compared to other tasks.

Next, we evaluate lifting the wine glass to assess the manipulation of large, fragile objects. While the glass's substantial size made grasping it straightforward, the task highlighted a significant difference in user confidence. Those operating the rigid hand exhibited noticeable hesitation, moving slowly to avoid applying excessive force that might shatter the glass, taking more than twice as long to complete the task. Conversely, users with \ourshort leveraged the hand's passive compliance to close the grasp rapidly, trusting the soft hand to compensate for contact force. Since the glass's substantial size made grasping straightforward, both hands achieved a 100\% success rate.

Handling the egg introduced the dual challenge of low surface friction and moderate fragility. The low friction proved to be the deciding factor, forcing LEAP users to operate with extreme caution and often taking significantly longer to attempt finding a stable grasp. \ourshort mitigated these risks through its compliant surface materials, which increased contact area, allowing users to secure the egg approximately 15 seconds faster on average than with the rigid hand. The rigid hand frequently failed to find a stable grasp before the time limit or slipped due to the lack of friction, resulting in a 60\% failure rate. \ourshort achieved a 90\% success rate.

\begin{figure*}[tp]
    \centering    \includegraphics[width=0.92\textwidth]{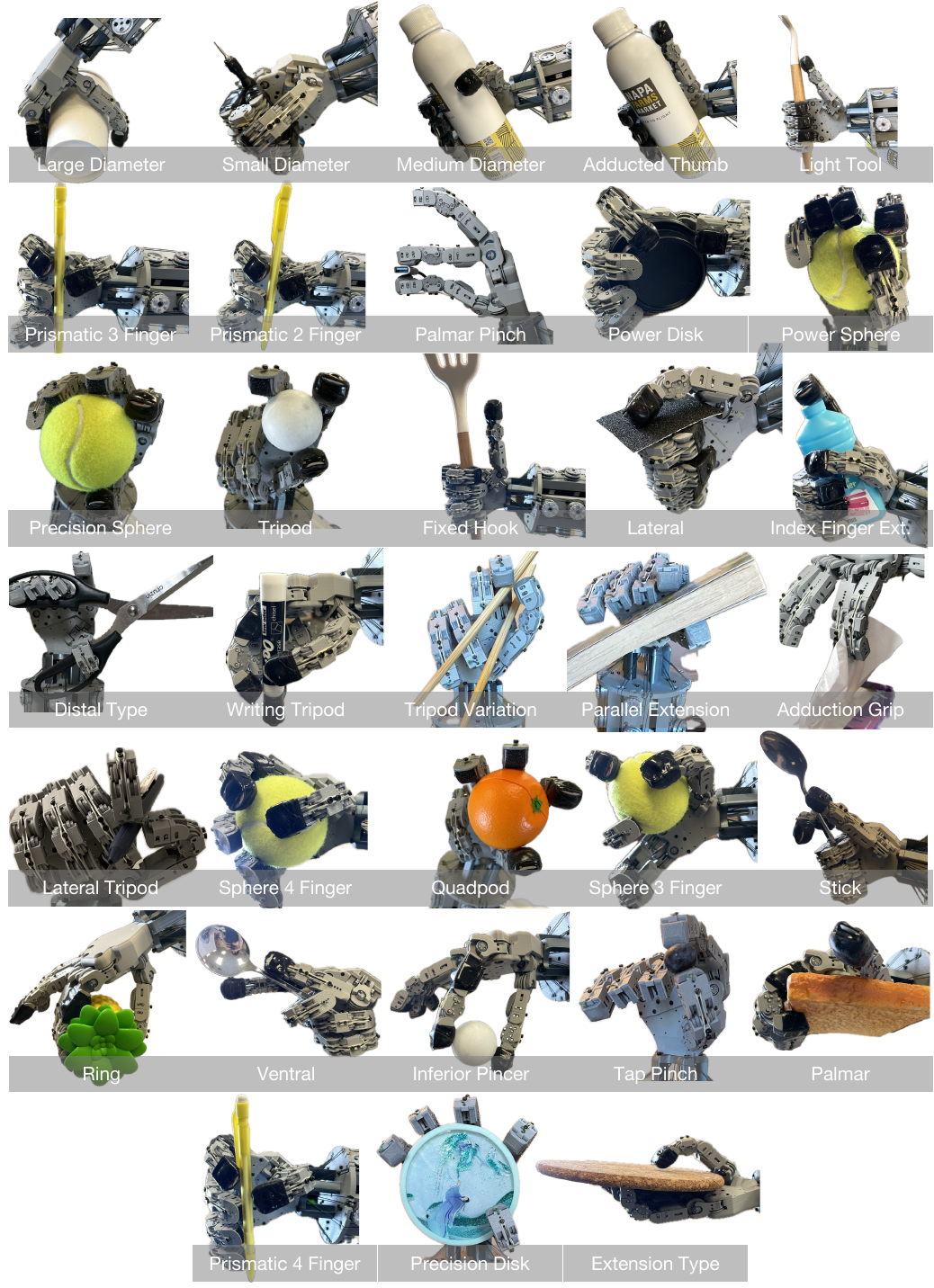}
    \caption{\small \textbf{Grasp Taxonomy Validation.} \ourshort demonstrating 33 distinct grasp types from the Feix taxonomy~\cite{feix2015grasp}. The hand successfully executes a wide range of configurations, including power grasps (e.g., Large Diameter), precision grasps (e.g., Prismatic Pinch), and intermediate poses, confirming that the compliant design maintains the kinematic versatility required for general-purpose manipulation.}
    \label{fig:grasp_poses}
\end{figure*}

Finally, picking the raspberry and the chip were extreme delicacy tests of small objects, requiring precise force application. While users still required significant time (approx. 40-50 s) to carefully align the grasp for these small targets, the task exposed a critical difference in force regulation. The rigid hand frequently crushed objects upon contact due to the lack of passive give, resulting in a 60\% success rate. \ourshort achieved a 100\% success rate on both tasks, as the soft fingers absorbed contact forces. \ourshort's compliance became the deciding factor.

\begin{comment}

\begin{table}[h]
    \centering
    \resizebox{0.8\linewidth}{!}{%
    \begin{tabular}{@{} l cc cc @{}}
        \toprule
        \textbf{Task} & \multicolumn{2}{c}{\textbf{Time Taken (s)} $\downarrow$} & \multicolumn{2}{c}{\textbf{Success Rate} $\uparrow$} \\
        \cmidrule(lr){2-3} \cmidrule(lr){4-5}
             & LEAP & \textbf{\our} & LEAP & \textbf{\our} \\
        \midrule
        Ball        & 24.1 & \textbf{7.4}  & 0.3 & \textbf{0.8} \\
        Wine Glass  & 10.9 & \textbf{4.3}  & \textbf{1.0} & \textbf{1.0} \\
        Egg         & 68.0 & \textbf{53.1} & 0.6 & \textbf{0.9} \\
        Raspberry   & 97.5 & \textbf{49.9} & 0.6 & \textbf{1.0} \\
        Chip        & 63.3 & \textbf{38.5} & 0.6 & \textbf{1.0} \\
        \bottomrule
    \end{tabular}
    }
    \caption{\small \textbf{Teleoperation User Study Results.} Comparison of average completion time and success rate for five manipulation tasks. \ourshort consistently outperforms the rigid LEAP hand, particularly in handling fragile objects (Raspberry, Chip) where compliance prevents damage, and in dynamic grasping (Ball) where adaptation simplifies capture.}
    \label{tab:teleop_results}
\end{table}

\end{comment}

\subsection{Grasp Taxonomy Assessment}
\label{subsec:grasp_taxonomy}

A crucial focus area in compliant robots is whether the introduction of compliance compromises the kinematic versatility required for general-purpose manipulation. To evaluate this, we assess \ourshort on the standardized benchmark of Grasp Taxonomy introduced by Feix et al.~\cite{feix2015grasp} and as attempted by previous works~\cite{zorin2025ruka}. This taxonomy formalizes human grasp types into a classification system, organizing grasps based on contact type, opposition direction, and hand shape. Consistent with prior works on open, robot hands~\cite{zorin2025ruka}, we evaluate on this benchmark because it represents the widest possible range of practically useful grasps required for daily activities. 

\begin{figure*}[t]
  \centering
  \includegraphics[width=0.9\linewidth]{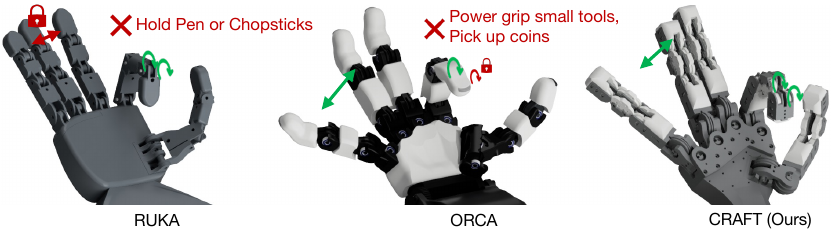}
  \vspace{-0.3cm} 
  \caption{\small \textbf{Finger Movement Comparison.} Unlike prior work, our hand incorporates both side-to-side movement from the base of the finger and active control of the first knuckle. In contrast, the RUKA hand \cite{zorin2025ruka} (RSS 2025) cannot move fingers sideways to squeeze objects, and the ORCA hand \cite{christoph2025orca} (IROS 2025) lacks active control of the fingertips. By utilizing 20 degrees of freedom, our design enables significantly higher dexterity for tasks like picking up coins or using pens/chopsticks.}
  \label{fig:dof_comparison}
  \vspace{-0.5cm}
\end{figure*}

\figref{fig:grasp_poses} reports \ourshort's performance on this standardized set of 33 grasp types, achieving 33/33 successes compared to RUKA's 29/33. Tendon-driven actuation provides the flexion torque to secure objects in power configurations, like the Large Diameter (row 1, col 1). Complementing this, soft material properties facilitate conformal contact for precision grasps, such as the Prismatic 2 Finger (row 2, col 2), stabilizing the hold through local deformation. We highlight performance on configurations typically challenging for compliant mechanisms. The Adduction Grip (row 4, col 5) demonstrates the index finger's ability to maintain lateral stiffness against the thumb’s orthogonal force. Similarly, the Writing Tripod (row 4, col 2) shows distal links can maintain specific geometry without collapsing under contact pressure. Finally, the Extension Type (row 7, col 3) validates the hand's full range of motion, forming a stable, flat platform for supporting objects.

\subsection{Whole Arm Teleopration}
\label{subsec:whole_arm_teleop}

\begin{figure}[t] 
    \vspace{-0.1cm}
  \centering
  \includegraphics[width=1\linewidth]{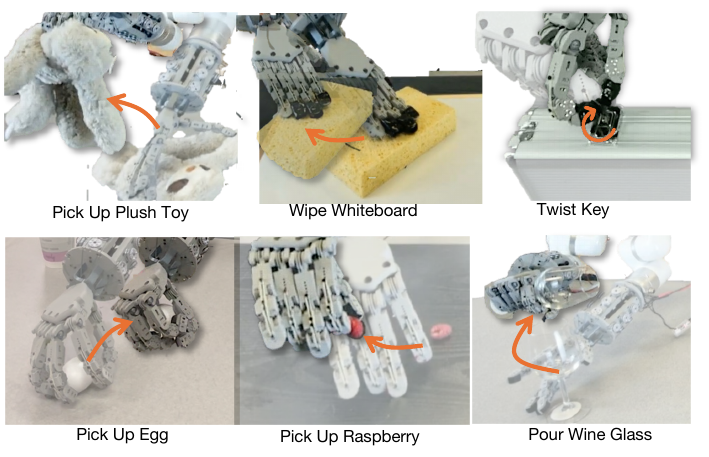}
  \caption{\small \textbf{Whole Arm Teleoperation.} \ourshort executing various tasks, including wiping, picking fragile objects, and twisting a key.}
  \label{fig:teleoperation}
  \vspace{-2mm}
\end{figure}

To evaluate the practical utility of \ourshort across diverse manipulation scenarios, we demonstrate teleoperation on open-world tasks shown in \figref{fig:teleoperation}. Picking up a plush toy validates grasping of soft objects. Wiping a whiteboard with a sponge demonstrates the benefit of compliance during sustained surface contact, as the soft joints maintain consistent pressure. Twisting a key illustrates the advantage of compliance in contact-rich scenarios, where the joints absorb reaction forces during torque. Picking up a raspberry tests precision grasping of small items, handling an egg requires managing low-friction fragility, and pouring from a wine glass validates stable manipulation of large breakable objects during dynamic motion. The hand completes all demonstrations without object damage or grasp failure, confirming that the hybrid design maintains functionality across a spectrum of functional tasks.

\subsection{Structural Comparison to Representative Robot Hands}
\label{subsec:structural_comparison}

\begin{figure}[t] 
  \centering
  \includegraphics[width=1\linewidth]{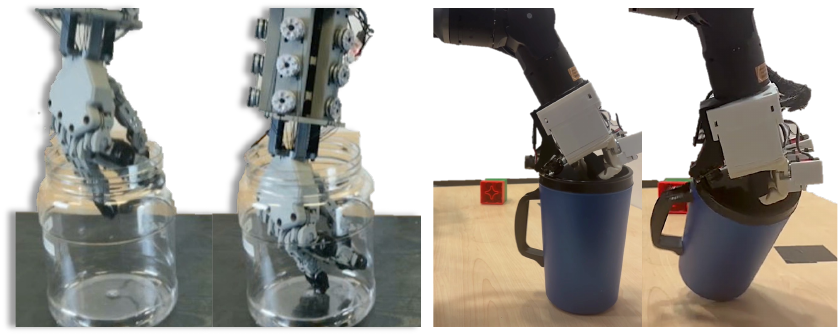}
  \caption{\small \textbf{Narrow Space Accessibility.} \ourshort demonstrates the ability to fit into a narrow jar, while the bulkier Leap Hand V2 fails to enter the same workspace due to its rigid linkage size. Both cans have the same radius. Results reproduced with the help of the authors of Leap Hand V2~\cite{shaw2025leap}.}
  \label{fig:narrow_reach}
  \vspace{-4mm}
\end{figure}

We compare the \ourshort design with representative anthropomorphic hands. While many other designs simplify kinematics to reduce complexity, \ourshort maintains full articulation within a compact form factor. We specifically contrast our kinematic structure with the RUKA and ORCA hands in \figref{fig:dof_comparison}, and evaluate our compact physical profile against the Leap Hand V2 in \figref{fig:narrow_reach}

\vspace{2mm}
\noindent\textbf{RUKA~\cite{zorin2025ruka}}: The RUKA hand utilizes a simplified structure that omits abduction/adduction DoF. While this reduces mechanical complexity, it completely eliminates the lateral workspace of the fingers. This limitation prevents inter-finger grasps, such as holding objects between fingers, and does not allow the manipulation of tools like chopsticks or pens, which rely on lateral forces between fingers. The RUKA hand is unable to execute four Feix grasps due to this constraint, whereas \ourshort achieves full taxonomy coverage.

\vspace{2mm}
\noindent\textbf{ORCA~\cite{christoph2025orca}}: Similarly, the ORCA hand simplifies structure by employing a fixed DIP joint. This reduces the effective curling range of the fingers and limits the hand's ability to adjust the fingertips necessary for precise tasks. This, again, limits its ability to perform tasks such as holding chopsticks and a pen.

\vspace{2mm}
\noindent\textbf{Leap V2~\cite{shaw2025leap}}: Leap V2 also adopts a hybrid hard-soft architecture but places some of its actuators behind the palm and fingers, resulting in a bulkier form factor than \ourshort. As demonstrated in \figref{fig:narrow_reach}, this limits its access in confined spaces: where Leap V2 collides with the environment, \ourshort's slim profile allows it to reach in.

\section{Conclusion}
We introduce \ourshort, an open-source anthropomorphic hand built for the reality of dexterous learning: lots of contact, repetition, and not much patience for broken hardware. \ourshort follows a hybrid hard-soft design, using rigid links to carry loads and compliant joints to absorb impacts. Rolling-contact joints at the PIP and DIP locations distribute load across the joint surface, preventing the stress concentrations that cause flexure joints to fracture under repeated use, and constrain motion to a well-defined kinematic path regardless of loading conditions. In structural benchmarks, \ourshort matches the repeatability of a rigid baseline while improving strength and reducing effort during sustained holding. In teleoperation, passive compliance reduces operator effort and contact failures, enabling reliable handling of fragile and low-friction items. Finally, \ourshort achieves full coverage of all 33 grasps in the Feix taxonomy, from power grasps to precision pinches. By releasing the full stack, we aim to make durable, repeatable dexterous data collection and policy learning easier to reproduce and scale.
\section{Acknowledgements}
We thank the Siebel School Robotics Lab at UIUC and the School of Information and Computer Sciences at UC Irvine for their support. We are grateful to Akash Sharma, Kenneth Shaw, and Yuchen Song for feedback on the draft. We also thank Steven Oh and Kenneth Shaw for assistance with LEAP V2 demonstrations, as well as the members of UIUC SIGRobotics for helpful discussions. This work was supported in part by the 1517 Fund and Dynamixel.

%% Use plainnat to work nicely with natbib. 

\bibliographystyle{plainnat}
\bibliography{references}

\end{document}